\documentclass[letterpaper, 10 pt, conference]{ieeeconf}  

\IEEEoverridecommandlockouts                              

\usepackage{graphics} 
\usepackage{epsfig} 
\usepackage{mathptmx} 
\usepackage{times} 
\usepackage{amsmath} 
\usepackage{amssymb}  
\usepackage{multirow}
\usepackage{hyperref}

\title{\LARGE \bf
Preparation of Papers for IEEE Sponsored Conferences \& Symposia*
}

\title{\LARGE \bf
Towards Cross-View-Consistent Self-Supervised Surround Depth Estimation}

\author{Laiyan Ding$^{1}$, Hualie Jiang$^{2}$, Jie Li$^{3}$, Yongquan Chen$^{4}$, and Rui Huang$^{1*}$
\thanks{This work was supported by Shenzhen Science and Technology Program under Grants JCYJ20220818103006012 and 20231128093642002, Guangdong Basic and Applied Basic Research Foundation under Grants 2023A1515011347 and 2023A1515110729, Longgang District supporting funds for Shenzhen ``Ten Action Plans" under Grant LGKCSDPT2024002, and Research Foundation of Shenzhen Polytechnic University under Grant 6023312007K.}
\thanks{$^{*}$Corresponding author}
\thanks{$^{1}$Laiyan Ding and Rui Huang are with School of Science and Engineering, The Chinese University of Hong Kong, Shenzhen 518172, China (e-mail: laiyanding@link.cuhk.edu.cn; ruihuang@cuhk.edu.cn).}
\thanks{$^{2}$Hualie Jiang is with Insta360 Research, Shenzhen 518000, China (e-mail: jianghualie@insta360.com).}
\thanks{$^{3}$Jie Li is with school of Artificial Intelligence, Shenzhen Polytechnic University, Shenzhen 518055, China (e-mail: jieli1@szpu.edu.cn).}
\thanks{$^{4}$Yongquan Chen is with Shenzhen Institute of Artificial Intelligence and Robotics for Society, The Chinese University of Hong Kong, Shenzhen 518172, China (e-mail: yqchen@cuhk.edu.cn).}}

\begin{document}

\maketitle
\thispagestyle{empty}
\pagestyle{empty}

\begin{abstract}
Depth estimation is a cornerstone for autonomous driving, yet acquiring per-pixel depth ground truth for supervised learning is challenging. Self-Supervised Surround Depth Estimation (SSSDE) from consecutive images offers an economical alternative. While previous SSSDE methods have proposed different mechanisms to fuse information across images, few of them explicitly consider the cross-view constraints, leading to inferior performance, particularly in overlapping regions. This paper proposes an efficient and consistent pose estimation design and two loss functions to enhance cross-view consistency for SSSDE. For pose estimation, we propose to use only front-view images to reduce training memory and sustain pose estimation consistency. The first loss function is the dense depth consistency loss, which penalizes the difference between predicted depths in overlapping regions. The second one is the multi-view reconstruction consistency loss, which aims to maintain consistency between reconstruction from spatial and spatial-temporal contexts. Additionally, we introduce a novel flipping augmentation to improve the performance further.
Our techniques enable a simple neural model to achieve state-of-the-art performance on the DDAD and nuScenes datasets. Last but not least, our proposed techniques can be easily applied to other methods. The code is available at \href{https://github.com/denyingmxd/CVCDepth}{https://github.com/denyingmxd/CVCDepth}.

\end{abstract}

\section{INTRODUCTION}

\label{sec:introduction}
Depth perception is a crucial component of reliable autonomous driving and robotics. Nevertheless, due to the high cost of deploying depth sensors, e.g., LiDAR, acquiring high-quality depth from images becomes an attractive alternative. Recent years have witnessed remarkable development of image-based depth estimation \cite{zhou2017unsupervised,chang2018pyramid,godard2019digging,ranftl2021vision,bhat2023zoedepth} and its applications in various scenarios, including 3D object detection \cite{li2023bevdepth,wang2022sts} and BEV segmentation \cite{philion2020lift,liu2022bevfusion}, etc.

In the field of image-based depth estimation, self-supervised depth estimation from images is of particular interest since it eliminates the need for depth supervision or stereo rectification. It utilizes the image reconstruction from temporal frames as supervision to train the depth and pose network jointly \cite{godard2019digging,guizilini20203d,jiang2020dipe,zhao2022monovit,zhang2023lite}. However, these methods can only infer scale-ambiguous depth \cite{zhou2017unsupervised}.

\begin{figure}[t]
    \centering
    \includegraphics[width=0.5\textwidth]{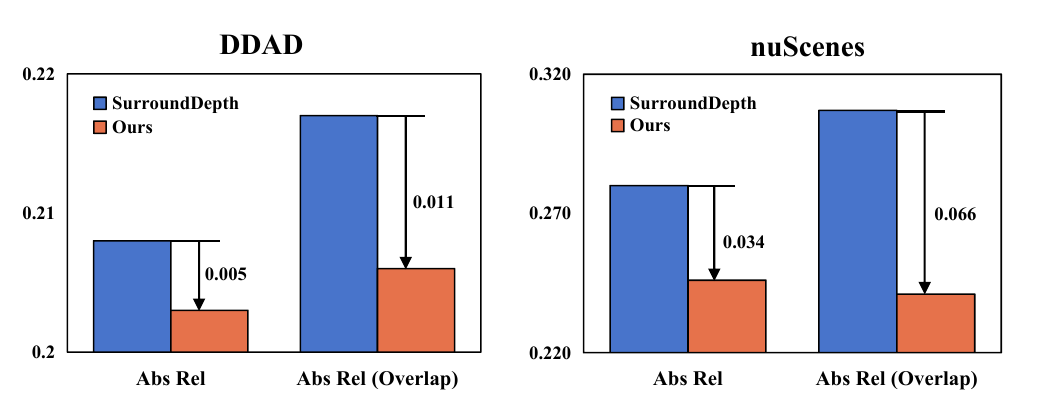}
    \caption{Performance comparison between ours and previous SOTA SurroundDepth \cite{wei2023surrounddepth} in all regions and overlapping regions.}
    \label{fig:comparison}
\end{figure}

Recently, self-supervised surround depth estimation has been proposed \cite{guizilini2022full}. This surround-view perception task takes advantage of the multi-camera setup in modern autonomous driving scenarios \cite{caesar2020nuscenes,guizilini20203d}. The scale-aware poses between cameras and overlapping regions among spatially neighbouring views can help recover scale-aware depth in self-supervised learning. Subsequently, various methods for fusing information among different views have been proposed to improve depth estimation accuracy \cite{wei2023surrounddepth, kim2022self, shi2023ega, xu2022multi}.

In this work, we propose an architectural design and novel losses to enhance cross-view consistency. Firstly, we only use front-view images for pose estimation and the relative camera poses to get the poses of other views. This design is motivated by the fact that front-view depth estimation is better than other views by large margins. Secondly, we introduce a novel dense depth consistency loss to penalize depth-prediction difference in overlapping regions, providing more dense and thus more effective supervision than the loss from MCDP \cite{xu2022multi}. Thirdly, we propose a loss function to penalize differences in image reconstruction from spatial and spatial-temporal contexts. To further boost the performance, a novel flipping technique for SSSDE is introduced. Current methods typically turn off the widely-used horizontal flipping augmentation since the flipping violates the camera relations. Nevertheless, we manage to apply flipping by modifying the training process carefully.

Together, these contributions yield state-of-the-art SSSDE results on the DDAD \cite{guizilini20203d} and nuScenes \cite{caesar2020nuscenes} datasets with only a simple model. As shown in Figure \ref{fig:comparison}, we achieve lower $Abs \ Rel$ compared to the previous SOTA SurroundDepth \cite{wei2023surrounddepth}, especially on the more challenging nuScenes\cite{caesar2020nuscenes} dataset. Moreover, the performance gains in overlapping regions are almost doubled, indicating that our method can produce more cross-view-consistent results.

\section{Related Work}
\begin{figure*}[h]
    \centering
    \includegraphics[width=1.0\textwidth]{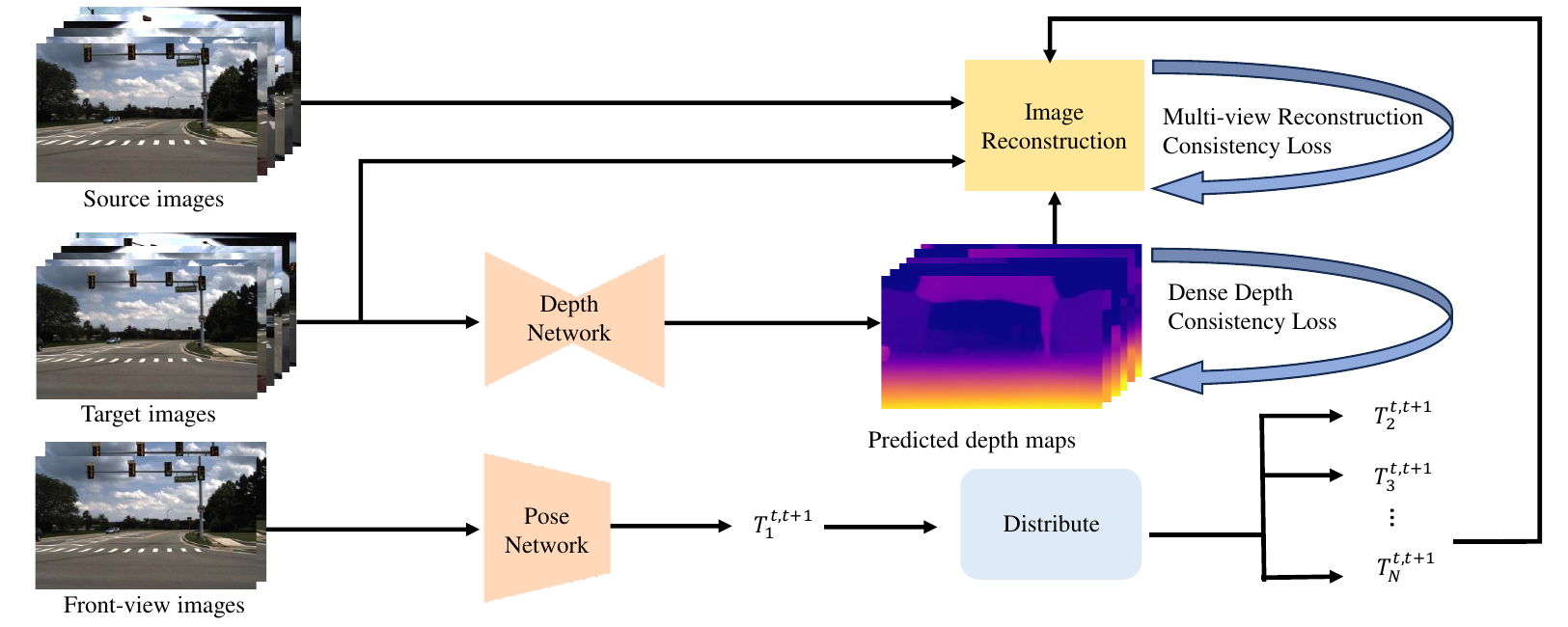}
    \caption{Overall pipeline of our proposed method. Augmentation is omitted for simplicity.}
    \label{fig:pipeline}
\end{figure*}
\subsection{Self-Supervised Monocular Depth Estimation} These approaches eliminate the need for ground truth and rectified stereo pairs for depth learning. Instead, they learn depth and motions simultaneously \cite{zhou2017unsupervised,casser2019depth,peng2021excavating,shu2020feature,zhao2022monovit,zhang2023lite}. The supervision signals come from the reconstruction error between the reference image and the reconstructed image from temporally neighbouring frames. Nevertheless, these methods can only produce scale-ambiguous depth \cite{zhou2017unsupervised}.

\subsection{Self-Supervised Surround Depth Estimation} FSM \cite{guizilini2022full} is the first work to introduce self-supervised surround depth estimation. It proposes spatial and spatial-temporal reconstruction to recover scale-aware depth. Furthermore, it designs a multi-camera pose consistency loss that penalizes pose-prediction differences among different views. Later, SurroundDepth \cite{wei2023surrounddepth} proposes to predict the pose of the vehicle and transform it back to poses in each view using camera extrinsics. Furthermore, a transformer-based Cross-View Transformer module is utilized to integrate features from different views \cite{vaswani2017attention}. VFDepth \cite{kim2022self} proposes a canonical pose estimation module that predicts the canonical pose of the front-view camera and distributes it to other views using camera extrinsics. Additionally, feature fusion among different views in 3D voxel space is conducted. MCDP \cite{xu2022multi} introduces a depth consistency loss and an iterative depth refinement method. More recently, R3D3 \cite{schmied2023r3d3} utilizes a complex SLAM system \cite{teed2021droid} and a refinement network to refine depth outputs from the SLAM system. Different from previous works, we propose two general loss functions to enhance the cross-view consistency of the SSSDE outputs.

\subsection{Data Augmentation}  Data augmentation is an effective solution to limited data and overfitting. Various augmentation techniques include geometric transformations, color space augmentations, mixing images, adversarial training, and generative adversarial networks \cite{shorten2019survey}. In the field of self-supervised depth estimation, most works \cite{zhang2023lite,lyu2021hr} follow MonoDepth2 \cite{godard2019digging} to apply color jittering and horizontal flipping as the training augmentation techniques. However, in the surround depth estimation setup, flipping is non-trivial as it would destroy the geometry relationship between cameras defined by camera extrinsics. In this work, we carefully exploit the widely used horizontal flipping augmentation for its potential in self-supervised surround depth estimation.

\section{Methodology}
In this section, we first review the self-supervised depth estimation. Then, we describe our overall architecture, followed by detailed descriptions of our pose estimation design, loss functions, and augmentation methods.

\subsection{Self-Supervised Depth Estimation}

Self-supervised monocular depth estimation aims to learn scale-ambiguous depth from a single image to bypass the high cost of collecting depth ground truth \cite{fu2018deep} for supervision and the need for rectified stereo image pairs \cite{chang2018pyramid}. Most of these approaches follow the pioneering work from Zhou \textit{et al.} \cite{zhou2017unsupervised}. The fundamental idea is to reconstruct the reference image with the source image, predicted depth and poses, and differentiable bilinear
sampling \cite{jaderberg2015spatial}. The commonly used photometric loss \cite{godard2019digging} is a weighted combination of Structural Similarity Index Measure (SSIM) \cite{wang2004image} and L1 loss. With this photometric supervision, depth and poses can be trained jointly in an end-to-end manner.

Self-Supervised Surround Depth Estimation (SSSDE) is an extension of self-supervised monocular depth estimation, first introduced by FSM \cite{guizilini2022full}. In addition to the temporal learning as in self-supervised monocular depth estimation, FSM \cite{guizilini2022full} also leverages the overlapping region in spatially neighbouring views. Using predicted depth and known camera extrinsics, neighbouring views can be partially reconstructed. This allows for the recovery of scale-aware depth since the extrinsics used here are in absolute scale. 

\subsection{Overview of Proposed Architecture}

Figure \ref{fig:pipeline} shows the entire network architecture, which includes depth and pose networks. Surround views are fed into the depth network to obtain depth predictions. The pose network takes only the front-view images and outputs the pose in the front-view. By utilizing known camera extrinsics, poses in other views can be recovered. Furthermore, the dense depth consistency loss is applied to predicted depth maps, and the multi-view reconstruction consistency loss is added in image reconstruction. Finally, we apply a novel augmentation method during training.

\subsection{Front-View Pose Only Design}

\begin{figure}[htbp]
    \centering
    \includegraphics[width=0.5\textwidth]{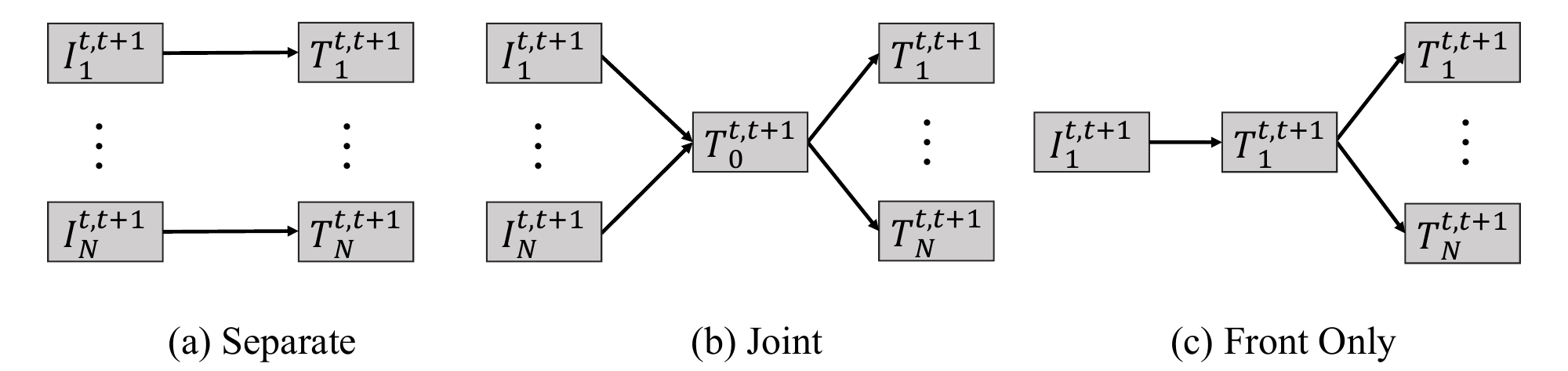}
    \caption{Comparison of different pose estimation methods.}
    \label{fig:pose_estimation}
\end{figure}

Pose estimation is a key component in self-supervised depth estimation. In the field of SSSDE, previous works have taken advantage of the fact that all cameras are attached to the vehicle and pose consistency among predictions for different views can be enforced. These methods can be grouped into two categories: (1) Separate pose prediction, e.g.,  FSM \cite{guizilini2022full} predicts the poses for different views separately and adds multi-camera pose consistency constraints. (2) Joint pose Prediction, e.g., SurroundDepth \cite{wei2023surrounddepth} and VFDepth \cite{kim2022self} combines the features from all views and decodes the fused features into the vehicle's pose or canonical pose, and poses for each view can be obtained using camera extrinsics. These approaches are depicted in Figure \ref{fig:pose_estimation} (a) and Figure \ref{fig:pose_estimation} (b).

\begin{table}[htbp]
\centering
\caption{Reproduced FSM \cite{guizilini2022full} results for each view on DDAD \cite{guizilini20203d}.}
\begin{tabular}{ |c|c|c|c|c|c|c| } 
\hline
Method &  Front & F.Left & F.Right & B.Left & B.Right & Back\\
\hline
FSM \cite{guizilini2022full} & 0.186 &  0.245 & 0.272 & 0.270 & 0.274 & 0.256 \\ \hline
\end{tabular}
\label{table:all_cams}
\end{table}

Unlike previous methods, we hypothesize that using only front-view images to regress the pose is already effective enough. As shown in Figure \ref{fig:pose_estimation} (c), we predict the front-view pose using only front-view images and then distribute it to other views. Our motivation is two-fold. First, front view depth estimation is significantly better than other views, as shown in Table \ref{table:all_cams}. Due to the tight link between ego-motion and depth predictions \cite{bian2019unsupervised}, we may assume the pose prediction in the front view is better than other views. Second, front-view information is sufficient to predict the front-view pose. Additional information in other views could barely help the front view pose regression. This can be validated by our experiments in Table \ref{table:poses}. Compared with other pose prediction methods, we only need one pass of encoding and decoding to process a batch of six surround views, while others need at least six passes of encoding, as seen from Figure \ref{fig:pose_estimation}. Consequently, our simple design can reduce memory consumption considerably during training. 

To be specific, taking pose prediction from time t to t+1 as an example, once we get the pose prediction $T_{1}^{t,t+1}$ for the front view, we distribute it to view i by leveraging the camera extrinsics ${E_i}$ as follows:

\begin{equation}
    T_i^{t, t+1}=E_i^{-1} E_1 T^{t, t+1}_{1} E_1^{-1} E_i
\end{equation}

\subsection{Dense Depth Consistency Loss}
\label{sec: ddcl}

Multi-view consistency is a common challenge in self-supervised depth estimation. SC-Depth  \cite{bian2019unsupervised} introduces temporal geometry consistency in monocular estimation and performs backward warping directly. However, MCDP \cite{xu2022multi} notices that due to the large difference in camera viewpoints, the depth maps estimated from spatially neighbouring cameras cannot be directly compared. Thus they propose Depth Consistency Loss (DCL) with forward warping, which leads to sparse supervision due to discretization. VFDepth \cite{kim2022self} maintains consistency of depth map at novel viewpoints. Yet, they require a 3D representation of the scene to synthesize the depth. Our idea shares the same insight as above to enforce depth consistency among different views.  Instead, we propose a Dense Depth Consistency Loss (DDCL) by transforming the source depth beforehand and performing backward warping later. Consequently, DDCL results both dense and correct supervision. Compared with previous methods, our DDCL is more effective and easy to apply for various architectures.

\begin{figure}[h]
    \centering
    \includegraphics[width=0.5\textwidth]{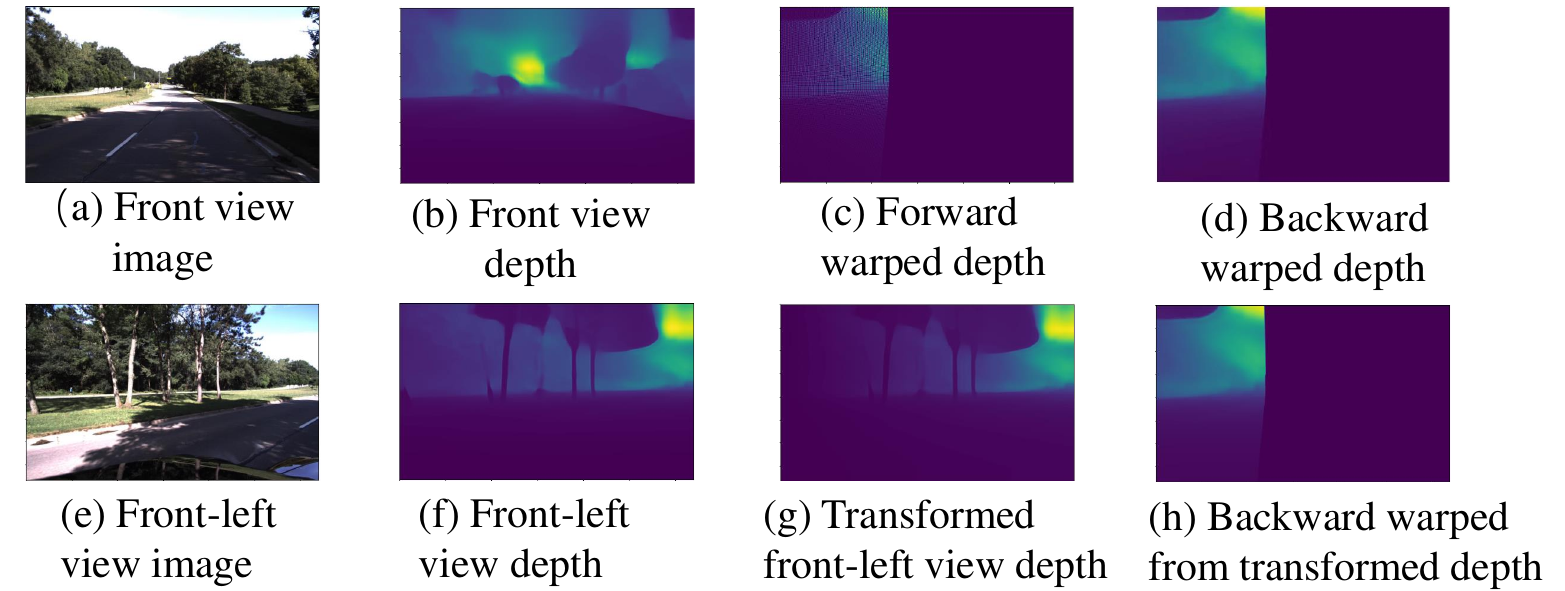}
    \caption{Different ways to project depth from spatially neighbouring views.}
    \label{fig:forward_backward_depth}
\end{figure}

As illustrated in Figure \ref{fig:forward_backward_depth}, taking the front and front-left images as an example, two depth maps are predicted. To project the depth of points from the front-left-view to the front-view, one may consider backward warping to generate a dense depth map (Figure \ref{fig:forward_backward_depth}d). However, due to the viewpoint difference, the same objects in the world coordinate would have different depths when projected into different camera views. Thus, backward warping is not the right way. One simple and correct way is to perform forward warping \cite{xu2022multi}. But, this will result in depth with holes due to discretization (Figure \ref{fig:forward_backward_depth}c). To overcome the above disadvantages, which may lead to suboptimal performance, we propose the novel DDCL. We first transform the front-left depth map by assuming it is in the front-view (Figure \ref{fig:forward_backward_depth}g). That is, for each point, we leverage its homogeneous coordinate, predicted depth, and extrinsics to compute its depth in the front-view. Then we perform backward warping to get the projected depth map (Figure \ref{fig:forward_backward_depth}h) with bilinear sampling \cite{zhou2017unsupervised}. Compared with forward warping \cite{xu2022multi}, more points in the target depth are supervised since there are no holes now.

For a surround view of $N$ cameras, where for camera $i$, the predicted depth map is $D_i$, the depth projected from neighbouring views using aforementioned operations is $\tilde{D_i}$, the DDCL is calculated as L1 loss:

\begin{equation}
    L_{DDCL} = \sum_{i=1}^{N}\left\|D_i-\tilde{D_i}\right\|_1
\end{equation}

\subsection{Multi-View Reconstruction Consistency Loss}

\begin{figure}[h]
    \centering
    \includegraphics[width=0.5\textwidth]{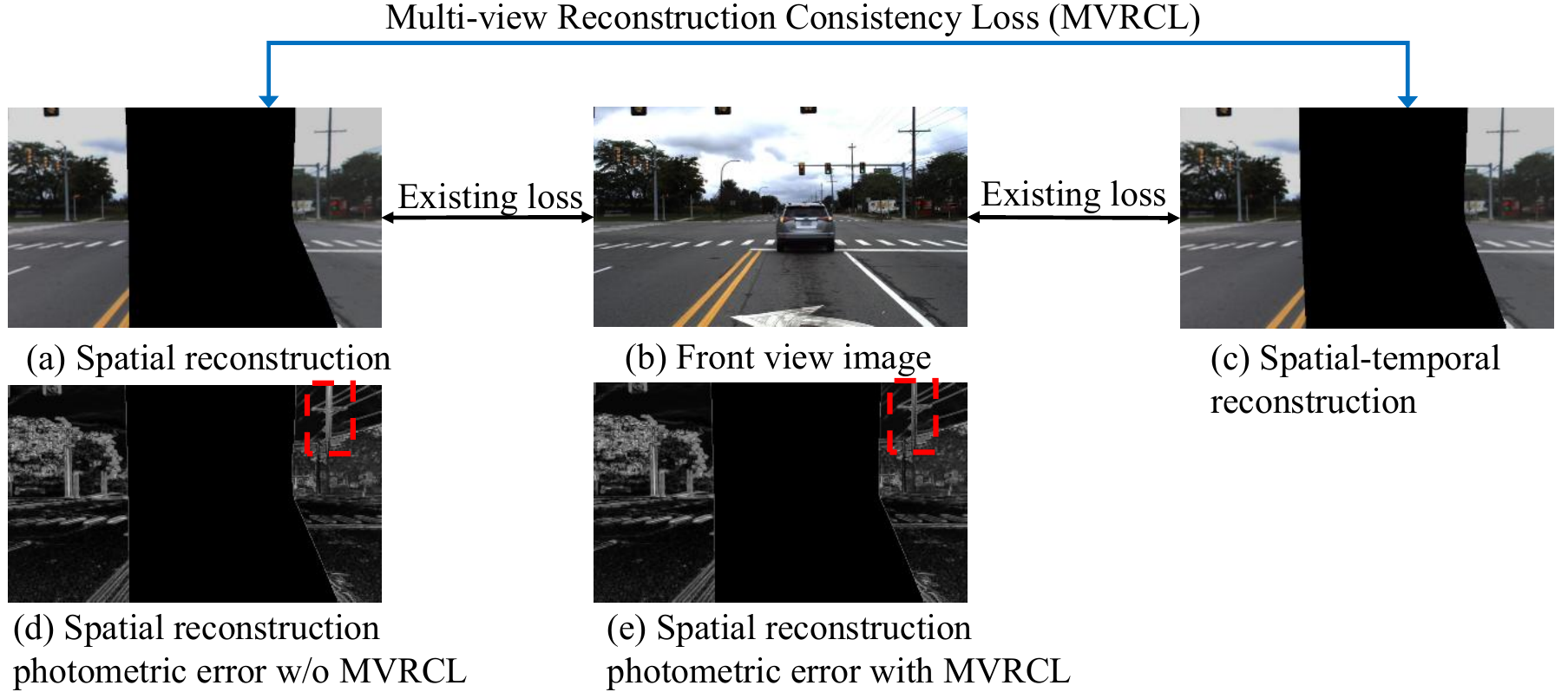}
    \caption{Our proposed multi-view reconstruction consistency loss can reduce spatial reconstruction errors.}
    \label{fig:sp_stp_con}
\end{figure}

Following FSM \cite{guizilini2022full}, we apply the spatial reconstruction to the recover metric scale and spatial-temporal reconstruction to further incorporate larger spatial-temporal contexts. A natural idea is to enforce the consistency between the reconstruction from spatial and spatial-temporal contexts. This can be achieved by adding one more reconstruction loss between the two reconstructed images, as illustrated in Figure \ref{fig:sp_stp_con}. Since this loss involves reconstructions from two views, i.e., the spatially neighbouring view and the spatial-temporally neighbouring view, it is named Multi-View Reconstruction Consistency Loss (MVRCL). From the perspective of the source image, this loss can be interpreted as maintaining temporal consistency under the target view.

Consequently, the spatial photometric error is reduced by comparing the red region in Figure \ref{fig:sp_stp_con} (d) and (e). This is similar to the phenomenon reported by FSM \cite{guizilini2022full}, where photometric error is reduced after applying spatial-temporal reconstruction. Our additional reconstruction consistency loss drives this photometric error from spatial contexts to be smaller.
Note that we do not enforce constraints between temporal and spatial or spatial-temporal reconstruction since images from different views can have quite different appearances due to large variations of viewpoint and changes of illuminance.

Given a surround view of $N$ cameras, for camera $i$, the original image is $I_i$, the reconstructed image from spatial contexts is $\tilde{I}^{s}_i$, the reconstructed image from spatial-temporal contexts is $\tilde{I}^{st}_i$, and the MVRCL is calculated as an image reconstruction loss:

\begin{equation}
    L_{MVRCL} = \sum_{i=1}^{N}(1-\alpha)\left\|\tilde{I}^s-\tilde{I}^{st}\right\|_1+\alpha \frac{1-\operatorname{SSIM}\left(\tilde{I}^s, \tilde{I}^{st}\right)}{2}
\end{equation}
where $\alpha$ is the weight of SSIM loss \cite{wang2004image}.

\subsection{Augmentation for Self-Supervised Surround Depth Estimation}

Augmentation is a crucial component to the success of deep learning methods \cite{shorten2019survey}. Self-supervised monocular depth estimation often utilizes horizontal flipping as a geometric augmentation \cite{godard2019digging}. However, in the field of SSSDE, horizontal flipping cannot be applied naively. The problem lies in the camera extrinsics, which describes the geometric relationship between different cameras. Flipping the images would destroy such relationships. 

To tackle this problem, we propose a novel \textbf{H}orizontal-flip augmentation for self-supervised \textbf{S}urround depth estimation (Hflip-S) that operates differently on depth network and pose network. The core idea is that outputs from the networks, with flipping applied to inputs, should be appropriately transformed as if the inputs were not flipped.  

\begin{figure}[h]
    \centering
    \includegraphics[width=0.5\textwidth]{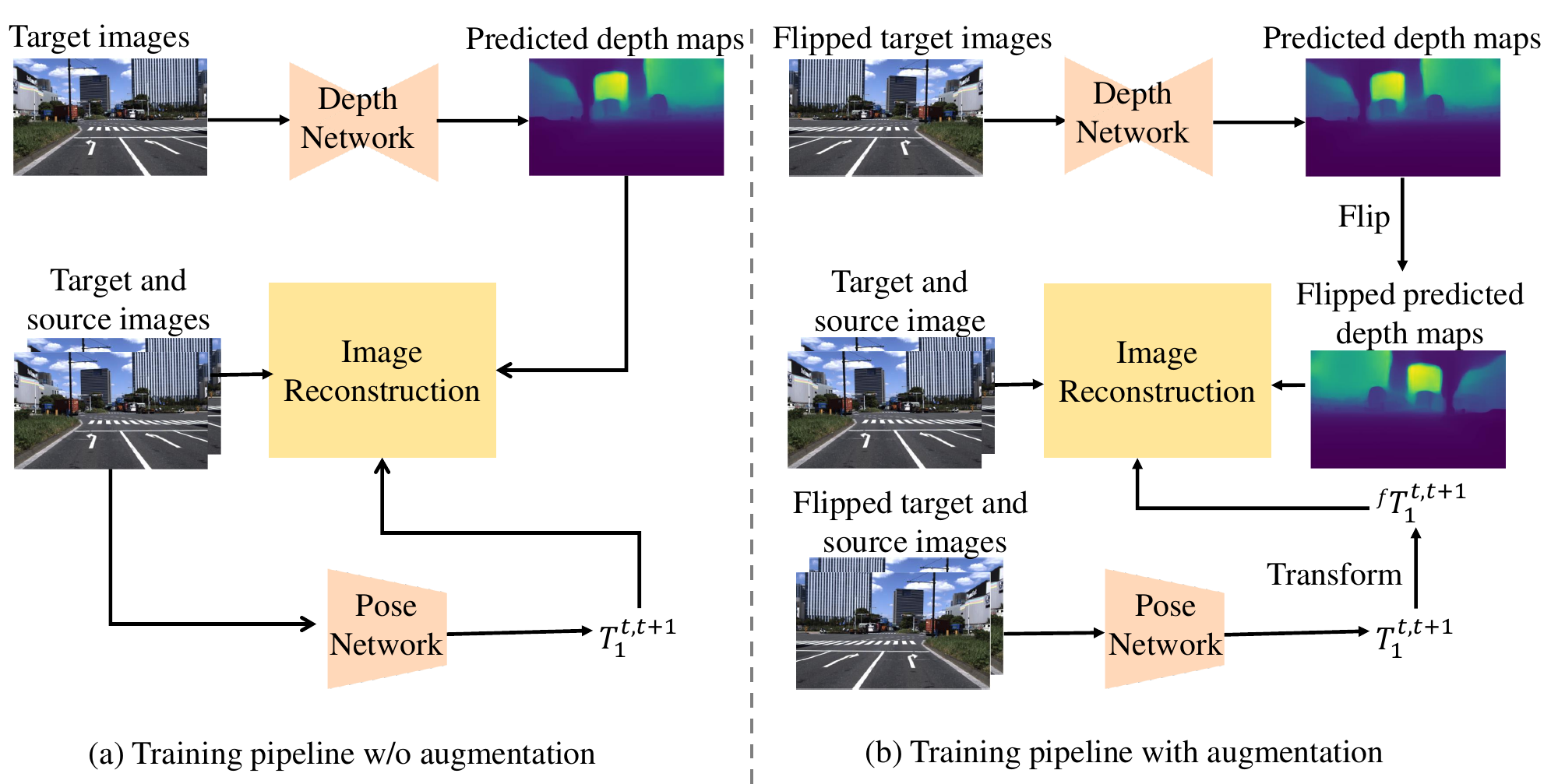}
    \caption{Comparison of training pipeline using augmentation or not.}
    \label{fig:aug}
\end{figure}

We apply the Hflip-S with a probability of 50\%. Figure \ref{fig:aug} shows the training process with or without flipping. When we apply horizontal flipping, the required operations for depth and pose prediction are different. For depth prediction, we input the flipped version of inputs to the network and flip the depth prediction back. For pose prediction, we transform the pose prediction for flipped inputs back using Eqn. \ref{eq:pose_transformation} from the appendix. By making modifications to the pipeline as shown in Figure \ref{fig:aug} (b), a novel augmentation for SSSDE is introduced. Furthermore, flipping for depth and pose networks can be used standalone, though we have validated that combining both can lead to better performance.

\section{Overall Training Loss}
Our overall training loss follows FSM \cite{guizilini2022full} with our proposed losses added:

\begin{equation}
\begin{aligned}
    L &= L_{t}+\lambda_{s}L_{s}+\lambda_{st}L_{st}+\lambda_{smooth}L_{smooth} \\
    &+\lambda_{DCCL}L_{DCCL}+\lambda_{MVRCL}L_{MVRCL}
\end{aligned}
    \label{all_loss}
\end{equation}
where $L_t$,$L_{s}$,$L_{st}$ are temporal, spatial, spatial-temporal photometric losses, and $L_{smooth}$ is the edge-aware smoothness.

\section{Experiments}

This section will first describe the datasets used and implementation details. We then compare our methods with other state-of-the-art methods quantitatively and qualitatively. Next, abundant ablation studies are performed to validate the effectiveness of proposed techniques. Lastly, we validate the versatility of our techniques on VFDepth \cite{kim2022self}.

\subsection{Datasets and Implementation Details}

Self-supervised surround depth estimation is evaluated on the multi-camera datasets DDAD \cite{guizilini20203d} and nuScenes \cite{caesar2020nuscenes}. Both datasets provide surround-view images, camera intrinsics and extrinsics, and depth ground truth for evaluation. The training resolution is of 384 $\times$ 640 and 352 $\times$ 640 for DDAD and nuScenes respectively, following VFDepth \cite{kim2022self}. The model is trained for 20 epochs on both datasets.

We implement our pipeline using Pytorch \cite{paszke2019pytorch} framework on four NVIDIA RTX 2080Ti GPUs. The network is trained with the following hyperparameters: Adam \cite{kingma2014adam} optimizer with $\beta_1 = 0.9$ and $\beta_2 = 0.999$, a batchsize of 6 per GPU (batchsize is 1 for each view), a learning rate of $10^{-4}$, with a StepLr scheduler that reduces the learning rate by $\frac{1}{10}$ at $\frac{3}{4}$ of the training epochs. For hyperparameters in Eqn. \ref{all_loss}, we set $\lambda_{s} = 0.03$, $\lambda_{st} = 0.1$, $\lambda_{smooth} = 0.1$, $\lambda_{DDCL} = 1 \times 10^{-3}$, $\lambda_{MVRCL} = 0.2$. Note that our proposed DDCL and MVRCL do not require extra estimation of the depth of adjacent frames. Thus, little overhead and memory consumption are introduced. Also, focal normalization \cite{facil2019cam} and intensity alignment \cite{kim2022self} are applied for stable training.

For evaluation, we follow previous works \cite{kim2022self,wei2023surrounddepth} to evaluate depth predictions up to 200m for DDAD \cite{guizilini20203d} and 80m for nuScenes \cite{caesar2020nuscenes}. Metrics from Eigen \textit{et al.} \cite{eigen2014depth} are adopted. We do not conduct post-processing \cite{godard2019digging} unless specified.

\subsection{Quantitative Experiments}

We compare our methods against other state-of-the-art methods in this section. Scale-aware and scale-ambiguous results are listed in Table \ref{tab:quantitative_scale} and Table \ref{tab:quantitative_ambiguous}, respectively. We also include the training memory using a batchsize of 6 (batchsize is 1 for each view) in Table \ref{tab:quantitative_scale}. Our baseline method is a reproduced variant of FSM \cite{guizilini2022full} where we use the pose network from MonoDepth2\cite{godard2019digging} following VFDepth \cite{kim2022self}. Note that scale-aware evaluation is more meaningful to real applications but more challenging.

For a fair comparison and easier understanding, the following symbols and acronyms are used in Table \ref{tab:quantitative_scale} and Table \ref{tab:quantitative_ambiguous}: (1) The symbol $\oplus$ and $\star$ denote results reproduced by VFDepth \cite{kim2022self} and us. (2) The symbol $\ddagger$ denotes entirely scale-ambiguous methods. (3) SurroundDepth-M and SurroundDepth-A are the scale-aware and scale-ambiguous models from SurroundDepth \cite{wei2023surrounddepth}. (4) pp means post-processing \cite{godard2019digging}. Newly added results that are different from the reported ones are obtained using the original public trained model and codebase under the common protocol.

\begin{table}[htbp]
 \caption{Scale-aware evaluation on DDAD \cite{guizilini20203d} and nuScenes datasets \cite{caesar2020nuscenes}. We report the average results from all views.  Best depth results among similar methods are \textbf{bolded}.}
 \label{tab:quantitative_scale}
\resizebox{\linewidth}{!}{
\begin{tabular}{llrrrrr} 
\hline
Dataset & Method &  Abs Rel$\downarrow$& Sq Rel$\downarrow$& RMSE$\downarrow$& $\delta<1.25$$\uparrow$  & Memory\\
\hline
\multirow{7}{*}{DDAD} 
& FSM \cite{guizilini2022full}  & \textbf{0.201} & - & - & - & -\\
& FSM$\oplus$ \cite{guizilini2022full} &0.228 & 4.409 & 13.433  & 0.687  & 7.2GB\\
& Our baseline (FSM$\star$\cite{guizilini2022full}) &0.252 & 4.382 &14.684 & 0.551 &  7.2GB\\
& VFDepth \cite{kim2022self}  & 0.218  & 3.660  & 13.327    & 0.674  & 15.9GB \\
& SurroundDepth-M \cite{wei2023surrounddepth} &0.208 & 3.371  & 12.977 & 0.693 & 19.7GB \\
& Ours & 0.210 & 3.458 & 12.876  & 0.704 & 7.0GB  \\
& Ours (res34+pp) & 0.203 & \textbf{3.363} & \textbf{12.805} & \textbf{0.706} & 7.6GB\\
\hline
\multirow{7}{*}{nuScenes}
& FSM \cite{guizilini2022full} & 0.297 & - & - & - & -\\
& FSM$\oplus$  \cite{guizilini2022full} &  0.319 & 7.534 & 7.860 & 0.716 & 6.8GB \\
& Our baseline (FSM$\star$\cite{guizilini2022full}) & 0.418 & 13.271 & 9.210 &  0.657 & 6.8GB \\
& VFDepth \cite{kim2022self}& 0.289 & 5.718 & 7.551 & 0.709 & 15.3GB\\
& SurroundDepth-M \cite{wei2023surrounddepth} & 0.280 & \textbf{4.401}& 7.467 &0.661 & 17.8GB\\
& Ours &0.264 & 5.525 & 7.178 & 0.763 & 6.4GB\\
& Ours (res34+pp) &\textbf{0.246} & 4.440& \textbf{6.789}& \textbf{0.764}& 7.2GB\\
\hline
\end{tabular}
}
\end{table}

 \begin{table}[htbp]
 \caption{Scale-ambiguous evaluation on DDAD \cite{guizilini20203d} and nuScenes datasets \cite{caesar2020nuscenes} with per-frame median scaling. We report the average results from all views. Best depth results among similar methods are \textbf{bolded}.}
\resizebox{\linewidth}{!}{ 
\begin{tabular}{llrrrr} 
\hline
Dataset & Method  & Abs Rel$\downarrow$ & Sq Rel$\downarrow$& RMSE$\downarrow$ & $\delta<1.25$$\uparrow$ \\
\hline
\multirow{10}{*}{DDAD} 
& EGA-Depth-LR$\ddagger$\cite{shi2023ega}& 0.195 & 3.211 & 12.117 & 0.743 \\
& EGA-Depth-HR$\ddagger$\cite{shi2023ega}  & \textbf{0.191} & 3.126 & \textbf{11.922} & 0.747 \\
& MCDP $\ddagger$ \cite{xu2022multi}& 0.193 & \textbf{3.111} & 12.264 & \textbf{0.811} \\
& SurroundDepth-A $\ddagger$\cite{wei2023surrounddepth}& 0.200 &3.392 &12.270 & 0.740\\
\cline{2-6}
& FSM \cite{guizilini2022full} & \textbf{0.202} &- & -&- \\
& FSM$\oplus$ \cite{guizilini2022full}& 0.219 &4.161 &13.163    &   0.703\\
& Our baseline (FSM$\star$\cite{guizilini2022full})& 0.239 & 4.648 & 13.461 &  0.671  \\
& VFDepth \cite{kim2022self} & 0.221& 3.549 & 13.031   &  0.681 \\
& SurroundDepth-M\cite{wei2023surrounddepth}& 0.205 &3.348 &12.641 & 0.716\\
& Ours & 0.208 & 3.380  &12.640 & 0.716\\
& Ours (res34+pp) & 0.204 & \textbf{3.327} & \textbf{12.489} & \textbf{0.720}\\ 
\hline
\multirow{10}{*}{nuScenes}
& EGA-Depth-LR$\ddagger$\cite{shi2023ega} & 0.239 & 2.357 & 6.801 & 0.723 \\
& EGA-Depth-HR$\ddagger$\cite{shi2023ega} & \textbf{0.223} & \textbf{1.987} & \textbf{6.599} & \textbf{0.732} \\
& MCDP \cite{xu2022multi} $\dagger$&0.237 & 3.030 & 6.822 &0.719 \\
& SurroundDepth-A $\ddagger$\cite{wei2023surrounddepth}& 0.245& 3.067& 6.835 & 0.719  \\
\cline{2-6}
& FSM \cite{guizilini2022full} & 0.299 &- &- &- \\
& FSM$\oplus$ \cite{guizilini2022full}&  0.301& 6.180 &7.892    & 0.729  \\
& Our baseline (FSM$\star$\cite{guizilini2022full}) & 0.374 & 10.243 & 8.754	& 0.696 \\
& VFDepth \cite{kim2022self}  & 0.271 &4.496 &7.391   &0.726 \\
& SurroundDepth-M\cite{wei2023surrounddepth}& 0.271& \textbf{3.749}& 7.279 & 0.681  \\
& Ours& 0.258 & 4.540 & 7.030& \textbf{0.756}\\
& Ours (res34+pp) & \textbf{0.247} & 3.791 & \textbf{6.704} &\textbf{0.756}\\
\hline
\end{tabular}}
\label{tab:quantitative_ambiguous}
\end{table}

\subsubsection{Scale-Aware Evaluation}
For scale-aware evaluation on the DDAD dataset, as shown in Table \ref{tab:quantitative_scale}, our method outperforms previous arts while using much less training memory. Compared with VFDepth \cite{kim2022self}, we achieve a boost of 0.008 on $Abs \ Rel$ metric using a resnet18 \cite{he2016deep} encoder. To compare with SurroundDepth-M \cite{wei2023surrounddepth}, we use a resnet34 \cite{he2016deep} encoder and conduct post-processing. We obtain better performance without multi-scale loss \cite{godard2019digging} and transformer-based fusion. Unfortunately, due to different implementation techniques and hyperparameter settings, both us and the authors from VFDepth \cite{kim2022self} cannot reproduce the original FSM \cite{guizilini2022full} results. SurroundDepth \cite{wei2023surrounddepth} even reports that they cannot recover the metric scale when the authors try to reproduce FSM \cite{guizilini2022full}. Nevertheless, our proposed techniques can boost our reproduced FSM \cite{guizilini2022full} by 0.042 on $Abs \ Rel$. 

For scale-aware evaluation on the more challenging nuScenes dataset, where images are taken under different weather conditions and times of day, and the overlapping regions among cameras are smaller, we achieve even greater performance gains. We obtain an improvement of 0.025 and 0.034 on the $Abs \ Rel$ metric compared with VFDepth  \cite{kim2022self} and SurroundDepth-M \cite{wei2023surrounddepth}, respectively. 

\subsubsection{Scale-Ambiguous Evaluation}
For scale-ambiguous evaluation, where the predicted depth is per-frame median-scaled \cite{zhou2017unsupervised}, the results are shown in Table \ref{tab:quantitative_ambiguous}. 

Compared with scale-aware methods from Table \ref{tab:quantitative_scale}, we still achieve better results. For example, on DDAD dataset, we get the best performance on three out of four depth metrics, i.e., $Abs \ Rel$, $RMSE$, $\delta < 1.25$. Furthermore, we outperform the original FSM \cite{guizilini2022full}, VFDepth, and SurroundDepth-M \cite{wei2023surrounddepth} by 0.052, 0.074, and 0.024 on $Abs \ Rel$ on the more challenging nuScenes dataset.

Also, note that SurroundDepth-A and SurroundDepth-M achieve a $Abs \ Rel$ of 0.245 and 0.271 on nuScenes \cite{caesar2020nuscenes}. This difference indicates that scale-aware SSSDE is harder to learn. Still, we obtain comparable results against EGA-Depth-LR \cite{shi2023ega} and MCDP \cite{xu2022multi}, which are entirely scale-ambiguous.


\subsection{Qualitative Results}
\begin{figure}[t!]
    \centering
    \includegraphics[width=0.5\textwidth]{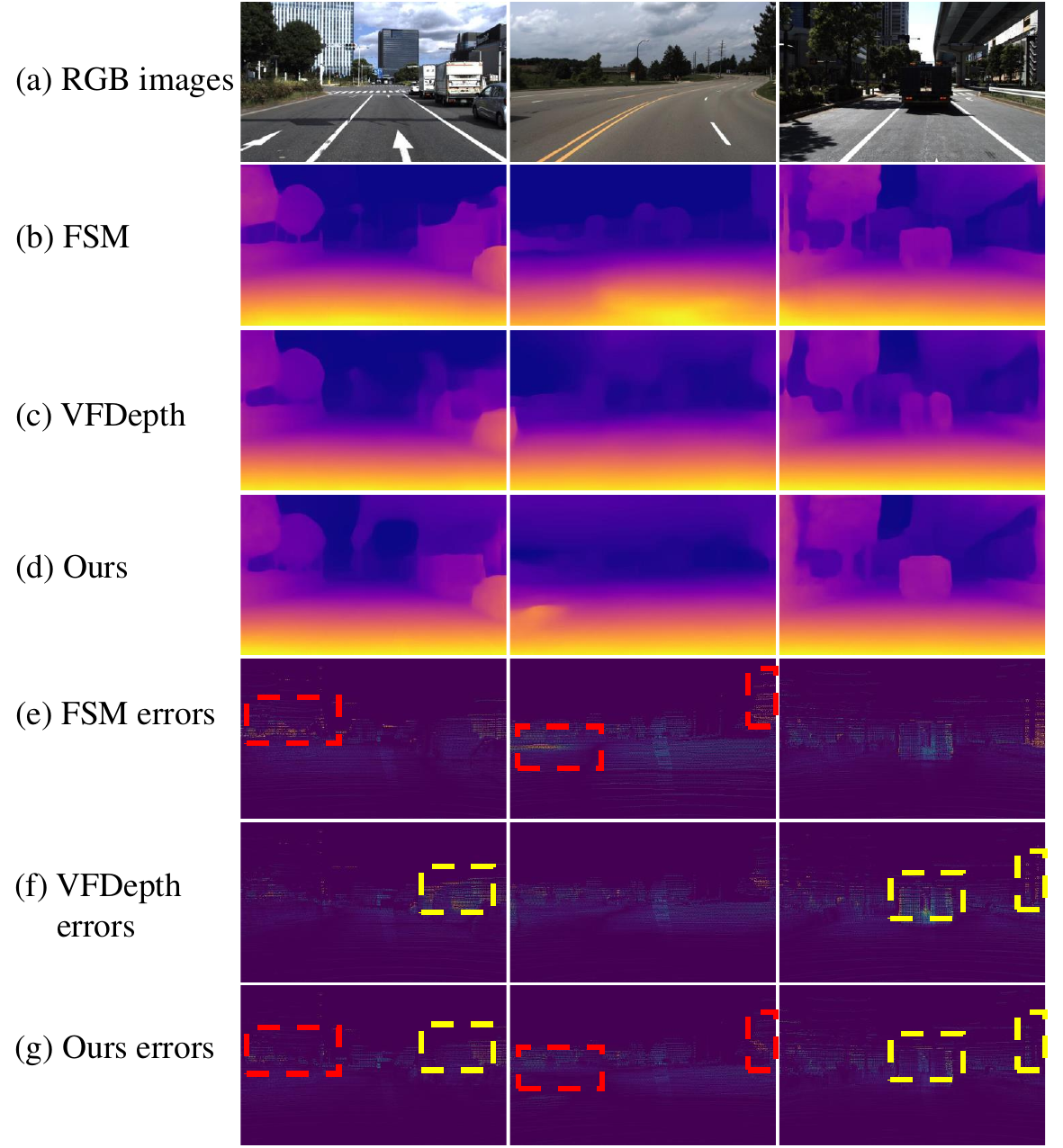}
    \caption{Qualitative comparisons among our reproduced FSM \cite{guizilini2022full} (our baseline), VFDepth\cite{kim2022self} and our full model on the DDAD dataset \cite{guizilini20203d}.}
    \label{fig:qualitative}
\end{figure}
We provide qualitative results to visualize the predictions of our methods and compare them with previous methods in Figure \ref{fig:qualitative}. The FSM \cite{guizilini2022full} and VFDepth \cite{kim2022self} results here are reproduced by us. Input images, predicted depth maps, and $Abs \ Rel$ error maps are included in Figure \ref{fig:qualitative}. The red rectangles indicate where our method outperforms our reproduced baseline FSM \cite{guizilini2022full}. For example, in the second column, with our proposed techniques applied, errors around the trees on the right side of the image are reduced, indicating that our techniques improve performance in overlapped regions. Furthermore, the yellow rectangles indicate where our method can outperform our reproduced VFDepth \cite{guizilini2022full}. For instance, in the third column, we achieve lower $Abs \ Rel$ on the car in the middle of the image and the building on the right of the image.

\subsection{Ablation Study}

In this section, we first provide an overall ablation study to verify the effectiveness of the proposed techniques. Then, we compare our proposed method with variants to validate the superiority of our method. Lastly, we apply our techniques to VFDepth \cite{kim2022self} to demonstrate versatility. All the experiments are conducted on DDAD \cite{guizilini20203d} dataset and scale-aware.

\subsubsection{Overall Ablation Study}

\begin{table}[h]
\centering
\caption{Ablation study on our proposed techniques.}
\resizebox{\linewidth}{!}{
\begin{tabular}{ |c|c|c|c|c|c|c| } 
\hline 
Front pose & DDCL       & MVRCL        &  Hflip-S    & Abs Rel$\downarrow$& Sq Rel$\downarrow$ & $\delta<1.25$$\uparrow$\\ \hline
    &     &   & &0.252 & 4.382 & 0.551 \\ \hline
\checkmark    &   &  &    & 0.229 & 4.361 & 0.676 \\ \hline
\checkmark    & \checkmark  &   &  & 0.215 & 3.634 & 0.693 \\ \hline
\checkmark    &   & \checkmark  &  & 0.224 & 4.397 &  0.697\\ \hline
\checkmark    &   &   & \checkmark &  0.222  & 4.182 & 0.702 \\ \hline
\checkmark    & \checkmark  & \checkmark  &  & 0.214 & 3.587 & 0.694 \\ \hline
\checkmark    & \checkmark  &   & \checkmark & 0.211 & 3.539 & 0.692 \\ \hline
\checkmark    & \checkmark  & \checkmark  & \checkmark & 0.208 &3.380 & 0.716 \\ \hline
\end{tabular}}
\label{table:ablation}
\end{table}

As shown in Table \ref{table:ablation}, all of our proposed techniques can improve the performance individually and jointly. Overall, we can obtain a large reduction on the $Abs \ Rel$ metric of 0.044. By comparing row 3, row 4, and row 5, we notice that DDCL can reduce both $Abs \ Rel$ and $Sq \ Rel$ greatly, while MVRCL and Hflip-S seem to improve more on the $\delta <1.25$ metric. This may indicate that DDCL has more impact on the farther points while MVRCL and Hflip-S improve more on the nearer points.

\subsubsection{Effectiveness of Front View Pose Only Design}

We verify that using front-view images and camera extrinsics is sufficient to acquire the poses for all views. We test four variants: (a) Pose Consistency: We use the pose consistency loss from FSM \cite{guizilini2022full}, which is also our baseline;  (b) Joint pose: We follow SurroundDepth \cite{wei2023surrounddepth} to extract features from all views, combine them and then decode them into the vehicle's poses; (c) Joint front pose: the same as (b) except the features are decoded into poses for the front-view. This is similar to VFdetph \cite{kim2022self}, except we conduct 2D fusions; (d) Front pose: only use the front-view images to get the poses of the front-view and distribute it to other views.

\begin{table}[htbp]
\centering
\caption{Comparisons between different pose estimation methods.}
\resizebox{\linewidth}{!}{
\begin{tabular}{ |c|c|c|c|c| } 
\hline
Method & \#Enc$\downarrow$ & \#Dec$\downarrow$ & Abs Rel$\downarrow$& Memory$\downarrow$ \\
\hline
Pose Consistency \cite{guizilini2022full} & 6 & 6 & 0.252& 7.227GB\\ \hline
Joint Pose \cite{wei2023surrounddepth}& 6 & 1 & 0.228 & 7.186GB\\ 
\hline
Joint Front Pose \cite{kim2022self}& 6 & 1 & 0.230 & 7.186GB \\ 
\hline
Front pose (Ours)& 1 & 1 & 0.229 & 6.113GB\\
\hline
\end{tabular}}
\label{table:poses}
\end{table}

From Table \ref{table:poses}, we can see that the front-view pose-only design requires one pass of encode and decode, greatly reducing the memory consumption during training. Furthermore, it achieves very similar results with joint pose or joint front pose. This verifies our intuition that using front-view images is sufficiently effective.

\subsubsection{Effectiveness of Dense Depth Consistency Loss}

As mentioned in Section \ref{sec: ddcl}, there exist two correct ways to implement the correct depth consistency loss: (1) DCL \cite{xu2022multi}: This can only provide sparse constraints on points that are projected. (2) DDCL: This applies transformation first and then uses backward warping to avoid holes. 


\begin{table}[htbp]
\centering
\caption{Effectiveness of our dense depth consistency loss.}
\begin{tabular}{ |c|c|c|c|c| } 
\hline
Method    & Abs Rel$\downarrow$& Sq Rel$\downarrow$  \\ \hline
w/o DC                   & 0.229 & 4.361 \\ \hline
DCL \cite{xu2022multi}          &  0.222   & 3.830 \\ \hline
DDCL & 0.215 & 3.634\\ \hline
\end{tabular}
\label{table:DC}
\end{table}

The comparisons in table \ref{table:DC} shows that both implementations can have a  performance boost. Nevertheless, our proposed DDCL can be more effective, benefiting from its dense supervision.

\begin{table}[htbp]
\centering
\caption{Comparisons between different augmentation methods.}
\begin{tabular}{ |c|c|c| } 
\hline
Method                  &  Abs Rel$\downarrow$ & Sq Rel$\downarrow$\\ \hline
w/o augmentation        & 0.229 & 4.361 \\ \hline
depth aug               & 0.225 & 4.354     \\ \hline
pose aug                & 0.227 & 4.362 \\ \hline
depth aug and pose aug  & 0.222  & 4.182 \\ \hline
\end{tabular}
\label{table:aug}
\end{table}

\subsection{Effectiveness of Our Proposed Augmentation}

As shown in Table \ref{table:aug}, augmentation for depth network and pose network can be effective alone compared with not using flipping augmentation. And, we find that using the two techniques together yields the best performance.

\subsection{Versatility of Proposed Techniques}

In this section, we choose VFDepth \cite{kim2022self} to validate the versatility of proposed techniques.
\begin{table}[h]
\centering
\caption{Applying our proposed techniques on VFdepth \cite{kim2022self}.}
\begin{tabular}{ |c|c|c|c|c| } 
\hline 
Front pose & DDCL       & MVRCL            & Abs Rel$\downarrow$ \\ \hline
           &          &    &0.231     \\ \hline
\checkmark            &          &               &0.230    \\ \hline
           & \checkmark   &      &   0.226   \\ \hline
           &          & \checkmark     &  0.222  \\ \hline
\checkmark           & \checkmark         & \checkmark     &  0.215  \\ \hline
\end{tabular}
\label{table:Versatility}
\end{table}
By comparing rows one and two, we again validate that the front pose design is effective and efficient. Furthermore, DDCL and MVRCL are still effective, though the network architecture between VFDepth \cite{kim2022self} and ours are quite different. By using these three techniques together, we achieve an improvement of 0.016 $Abs \ Rel$ against the reproduced VFDepth \cite{kim2022self}. However, it is not easy to apply our augmentation techniques since VFDepth would project features into 3D space, thus prohibiting flipping of the input images. So, we omit the experiment of using Hflip-S.

\section{Conclusion and Future Works}


In this paper, we have presented a simple model for SSSDE. We introduced four contributions which enable a simple model to achieve superior performance. Nevertheless, some limitations still exist, which may be handled in future works. First, though we have proposed several ways to maintain cross-view consistency, we do not conduct cross-view feature fusions. It is possible to apply techniques from SurroundDepth \cite{wei2023surrounddepth} and MCDP \cite{xu2022multi} to enhance the model further. Second, it is possible to apply techniques from MVSNet \cite{yao2018mvsnet}, including 3D cost volume building to obtain better performance.

\renewcommand{\theequation}{A.\arabic{equation}}

\setcounter{equation}{0}

\section*{Appendix}

Here, we provide a proof of conversion between the relative motions of target and source views, and their horizontally flipped counterparts, i.e., conversion between $T_t^s$ and ${}^{f}T_t^s$. 

Suppose that the camera corresponding to the target view is at the world origin, then coordinate transformation from the world coordinate $(X_w, Y_w, Z_w)$ to the source image coordinate $(u, v)$ can be expressed as, 

\begin{equation}
\begin{bmatrix}
u \\
v \\
1
\end{bmatrix} 
\simeq
\underbrace{\begin{bmatrix}
f_{x} & 0 & c_{x} \\
0 & f_{y} & c_{y} \\
0 & 0 & 1 
\end{bmatrix}}_{K}
\underbrace{
\begin{bmatrix}
r_{11} & r_{12} & r_{13} & t_1\\
r_{21} & r_{22} & r_{23} & t_2\\
r_{31} & r_{32} & r_{33} & t_3
\end{bmatrix}}_{T_t^s}
\begin{bmatrix}
X_w \\
Y_w \\
Z_w \\
1
\end{bmatrix}
\label{eq:proj}
\end{equation}

When the image of size $h \times w$ is horizontally flipped, the following transformation between the new point on image $(u', v')$ and the original point $(u, v)$ holds:
\begin{equation}
\begin{bmatrix}
u' \\
v' \\
1
\end{bmatrix} 
=
\begin{bmatrix}
w-u \\
v \\
1
\end{bmatrix} 
=
\begin{bmatrix}
-1 & 0 & w \\
0 & 1 & 0 \\
0 & 0 & 1 
\end{bmatrix}
\begin{bmatrix}
u \\
v \\
1
\end{bmatrix} 
\label{eq:flip_cam}
\end{equation}

By combining Eqn. \ref{eq:proj} and Eqn. \ref{eq:flip_cam}, we can get:

\begin{equation}
\begin{bmatrix}
u' \\
v' \\
1
\end{bmatrix} 
\simeq
\setlength\arraycolsep{2pt}
\underbrace{\begin{bmatrix}
f_{x} & 0 & w-c_{x} \\
0 & f_{y} & c_{y} \\
0 & 0 & 1 
\end{bmatrix}}_{K'}
\setlength\arraycolsep{2pt}
\underbrace{\begin{bmatrix}
r_{11} & -r_{12} & -r_{13} & -t_1\\
-r_{21} & r_{22} & r_{23} & t_2\\
-r_{31} & r_{32} & r_{33} & t_3
\end{bmatrix}}_{{}^fT_t^s}
\begin{bmatrix}
-X_w \\
Y_w \\
Z_w \\
1
\end{bmatrix} 
\quad \label{eq:flip2}
\end{equation}

We assume that the principle point is at the center of the image, so we can ignore the difference between $K$ and $K'$. 

Furthermore, after flipping, the world coordinate $(X_w', Y_w', Z_w')$ has the following relationship with the original world coordinate:
\begin{equation}
    \begin{bmatrix}
    X_w' \\
    Y_w' \\
    Z_w' \\
    1
    \end{bmatrix}
    =
    \begin{bmatrix}
    -X_w' \\
    Y_w' \\
    Z_w' \\
    1
    \end{bmatrix} 
    \label{eq: world_relation}
\end{equation}

From  Eqn. \ref{eq:proj}, Eqn. \ref{eq:flip2} and Eqn. \ref{eq: world_relation}, we can see that the relative motion of two views ($T_t^s$) and their horizontally flipped counterparts (${}^fT_t^s$) has the following relationship :
\begin{equation}
\setlength\arraycolsep{2pt}
\underbrace{\begin{bmatrix}
r_{11} & r_{12} & r_{13} & t_1\\
r_{21} & r_{22} & r_{23} & t_2\\
r_{31} & r_{32} & r_{33} & t_3
\end{bmatrix}}_{T_t^s} 
\leftrightarrow
\setlength\arraycolsep{2pt}
\underbrace{\begin{bmatrix}
r_{11} & -r_{12} & -r_{13} & -t_1\\
-r_{21} & r_{22} & r_{23} & t_2\\
-r_{31} & r_{32} & r_{33} & t_3
\end{bmatrix}}_{{}^fT_t^s}  
\label{eq:pose_transformation}
\end{equation}

\bibliographystyle{IEEEtran}
\bibliography{IEEEfull}

\begin{thebibliography}{10}
\providecommand{\url}[1]{#1}
\csname url@rmstyle\endcsname
\providecommand{\newblock}{\relax}
\providecommand{\bibinfo}[2]{#2}
\providecommand\BIBentrySTDinterwordspacing{\spaceskip=0pt\relax}
\providecommand\BIBentryALTinterwordstretchfactor{4}
\providecommand\BIBentryALTinterwordspacing{\spaceskip=\fontdimen2\font plus
\BIBentryALTinterwordstretchfactor\fontdimen3\font minus \fontdimen4\font\relax}
\providecommand\BIBforeignlanguage[2]{{%
\expandafter\ifx\csname l@#1\endcsname\relax
\typeout{** WARNING: IEEEtran.bst: No hyphenation pattern has been}%
\typeout{** loaded for the language `#1'. Using the pattern for}%
\typeout{** the default language instead.}%
\else
\language=\csname l@#1\endcsname
\fi
#2}}

\bibitem{zhou2017unsupervised}
T.~Zhou, M.~Brown, N.~Snavely, and D.~G. Lowe, ``Unsupervised learning of depth and ego-motion from video,'' in \emph{Proceedings of the IEEE conference on computer vision and pattern recognition}, 2017, pp. 1851--1858.

\bibitem{chang2018pyramid}
J.-R. Chang and Y.-S. Chen, ``Pyramid stereo matching network,'' in \emph{Proceedings of the IEEE conference on computer vision and pattern recognition}, 2018, pp. 5410--5418.

\bibitem{godard2019digging}
C.~Godard, O.~Mac~Aodha, M.~Firman, and G.~J. Brostow, ``Digging into self-supervised monocular depth estimation,'' in \emph{Proceedings of the IEEE/CVF international conference on computer vision}, 2019, pp. 3828--3838.

\bibitem{ranftl2021vision}
R.~Ranftl, A.~Bochkovskiy, and V.~Koltun, ``Vision transformers for dense prediction,'' in \emph{Proceedings of the IEEE/CVF international conference on computer vision}, 2021, pp. 12\,179--12\,188.

\bibitem{bhat2023zoedepth}
S.~F. Bhat, R.~Birkl, D.~Wofk, P.~Wonka, and M.~M{\"u}ller, ``Zoedepth: Zero-shot transfer by combining relative and metric depth,'' \emph{arXiv preprint arXiv:2302.12288}, 2023.

\bibitem{li2023bevdepth}
Y.~Li, Z.~Ge, G.~Yu, J.~Yang, Z.~Wang, Y.~Shi, J.~Sun, and Z.~Li, ``Bevdepth: Acquisition of reliable depth for multi-view 3d object detection,'' in \emph{Proceedings of the AAAI Conference on Artificial Intelligence}, vol.~37, no.~2, 2023, pp. 1477--1485.

\bibitem{wang2022sts}
Z.~Wang, C.~Min, Z.~Ge, Y.~Li, Z.~Li, H.~Yang, and D.~Huang, ``Sts: Surround-view temporal stereo for multi-view 3d detection,'' \emph{arXiv preprint arXiv:2208.10145}, 2022.

\bibitem{philion2020lift}
J.~Philion and S.~Fidler, ``Lift, splat, shoot: Encoding images from arbitrary camera rigs by implicitly unprojecting to 3d,'' in \emph{Computer Vision--ECCV 2020: 16th European Conference, Glasgow, UK, August 23--28, 2020, Proceedings, Part XIV 16}.\hskip 1em plus 0.5em minus 0.4em\relax Springer, 2020, pp. 194--210.

\bibitem{liu2022bevfusion}
Z.~Liu, H.~Tang, A.~Amini, X.~Yang, H.~Mao, D.~Rus, and S.~Han, ``Bevfusion: Multi-task multi-sensor fusion with unified bird's-eye view representation,'' in \emph{IEEE International Conference on Robotics and Automation (ICRA)}, 2023.

\bibitem{guizilini20203d}
V.~Guizilini, R.~Ambrus, S.~Pillai, A.~Raventos, and A.~Gaidon, ``3d packing for self-supervised monocular depth estimation,'' in \emph{Proceedings of the IEEE/CVF conference on computer vision and pattern recognition}, 2020, pp. 2485--2494.

\bibitem{jiang2020dipe}
H.~Jiang, L.~Ding, Z.~Sun, and R.~Huang, ``Dipe: Deeper into photometric errors for unsupervised learning of depth and ego-motion from monocular videos,'' in \emph{2020 IEEE/RSJ International Conference on Intelligent Robots and Systems (IROS)}.\hskip 1em plus 0.5em minus 0.4em\relax IEEE, 2020, pp. 10\,061--10\,067.

\bibitem{zhao2022monovit}
C.~Zhao, Y.~Zhang, M.~Poggi, F.~Tosi, X.~Guo, Z.~Zhu, G.~Huang, Y.~Tang, and S.~Mattoccia, ``Monovit: Self-supervised monocular depth estimation with a vision transformer,'' in \emph{2022 International Conference on 3D Vision (3DV)}.\hskip 1em plus 0.5em minus 0.4em\relax IEEE, 2022, pp. 668--678.

\bibitem{zhang2023lite}
N.~Zhang, F.~Nex, G.~Vosselman, and N.~Kerle, ``Lite-mono: A lightweight cnn and transformer architecture for self-supervised monocular depth estimation,'' in \emph{Proceedings of the IEEE/CVF Conference on Computer Vision and Pattern Recognition}, 2023, pp. 18\,537--18\,546.

\bibitem{wei2023surrounddepth}
Y.~Wei, L.~Zhao, W.~Zheng, Z.~Zhu, Y.~Rao, G.~Huang, J.~Lu, and J.~Zhou, ``Surrounddepth: Entangling surrounding views for self-supervised multi-camera depth estimation,'' in \emph{Conference on Robot Learning}.\hskip 1em plus 0.5em minus 0.4em\relax PMLR, 2023, pp. 539--549.

\bibitem{guizilini2022full}
V.~Guizilini, I.~Vasiljevic, R.~Ambrus, G.~Shakhnarovich, and A.~Gaidon, ``Full surround monodepth from multiple cameras,'' \emph{IEEE Robotics and Automation Letters}, vol.~7, no.~2, pp. 5397--5404, 2022.

\bibitem{caesar2020nuscenes}
H.~Caesar, V.~Bankiti, A.~H. Lang, S.~Vora, V.~E. Liong, Q.~Xu, A.~Krishnan, Y.~Pan, G.~Baldan, and O.~Beijbom, ``nuscenes: A multimodal dataset for autonomous driving,'' in \emph{Proceedings of the IEEE/CVF conference on computer vision and pattern recognition}, 2020, pp. 11\,621--11\,631.

\bibitem{kim2022self}
J.-H. Kim, J.~Hur, T.~P. Nguyen, and S.-G. Jeong, ``Self-supervised surround-view depth estimation with volumetric feature fusion,'' \emph{Advances in Neural Information Processing Systems}, vol.~35, pp. 4032--4045, 2022.

\bibitem{shi2023ega}
Y.~Shi, H.~Cai, A.~Ansari, and F.~Porikli, ``Ega-depth: Efficient guided attention for self-supervised multi-camera depth estimation,'' in \emph{Proceedings of the IEEE/CVF Conference on Computer Vision and Pattern Recognition}, 2023, pp. 119--129.

\bibitem{xu2022multi}
J.~Xu, X.~Liu, Y.~Bai, J.~Jiang, K.~Wang, X.~Chen, and X.~Ji, ``Multi-camera collaborative depth prediction via consistent structure estimation,'' in \emph{Proceedings of the 30th ACM International Conference on Multimedia}, 2022, pp. 2730--2738.

\bibitem{casser2019depth}
V.~Casser, S.~Pirk, R.~Mahjourian, and A.~Angelova, ``Depth prediction without the sensors: Leveraging structure for unsupervised learning from monocular videos,'' in \emph{Proceedings of the AAAI conference on artificial intelligence}, vol.~33, no.~01, 2019, pp. 8001--8008.

\bibitem{peng2021excavating}
R.~Peng, R.~Wang, Y.~Lai, L.~Tang, and Y.~Cai, ``Excavating the potential capacity of self-supervised monocular depth estimation,'' in \emph{Proceedings of the IEEE/CVF International Conference on Computer Vision}, 2021, pp. 15\,560--15\,569.

\bibitem{shu2020feature}
C.~Shu, K.~Yu, Z.~Duan, and K.~Yang, ``Feature-metric loss for self-supervised learning of depth and egomotion,'' in \emph{European Conference on Computer Vision}.\hskip 1em plus 0.5em minus 0.4em\relax Springer, 2020, pp. 572--588.

\bibitem{vaswani2017attention}
A.~Vaswani, N.~Shazeer, N.~Parmar, J.~Uszkoreit, L.~Jones, A.~N. Gomez, {\L}.~Kaiser, and I.~Polosukhin, ``Attention is all you need,'' \emph{Advances in neural information processing systems}, vol.~30, 2017.

\bibitem{schmied2023r3d3}
A.~Schmied, T.~Fischer, M.~Danelljan, M.~Pollefeys, and F.~Yu, ``R3d3: Dense 3d reconstruction of dynamic scenes from multiple cameras,'' in \emph{Proceedings of the IEEE/CVF International Conference on Computer Vision}, 2023, pp. 3216--3226.

\bibitem{teed2021droid}
Z.~Teed and J.~Deng, ``Droid-slam: Deep visual slam for monocular, stereo, and rgb-d cameras,'' \emph{Advances in neural information processing systems}, vol.~34, pp. 16\,558--16\,569, 2021.

\bibitem{shorten2019survey}
C.~Shorten and T.~M. Khoshgoftaar, ``A survey on image data augmentation for deep learning,'' \emph{Journal of big data}, vol.~6, no.~1, pp. 1--48, 2019.

\bibitem{lyu2021hr}
X.~Lyu, L.~Liu, M.~Wang, X.~Kong, L.~Liu, Y.~Liu, X.~Chen, and Y.~Yuan, ``Hr-depth: High resolution self-supervised monocular depth estimation,'' in \emph{Proceedings of the AAAI Conference on Artificial Intelligence}, vol.~35, no.~3, 2021, pp. 2294--2301.

\bibitem{fu2018deep}
H.~Fu, M.~Gong, C.~Wang, K.~Batmanghelich, and D.~Tao, ``Deep ordinal regression network for monocular depth estimation,'' in \emph{Proceedings of the IEEE conference on computer vision and pattern recognition}, 2018, pp. 2002--2011.

\bibitem{jaderberg2015spatial}
M.~Jaderberg, K.~Simonyan, A.~Zisserman, \emph{et~al.}, ``Spatial transformer networks,'' \emph{Advances in neural information processing systems}, vol.~28, 2015.

\bibitem{wang2004image}
Z.~Wang, A.~C. Bovik, H.~R. Sheikh, and E.~P. Simoncelli, ``Image quality assessment: from error visibility to structural similarity,'' \emph{IEEE transactions on image processing}, vol.~13, no.~4, pp. 600--612, 2004.

\bibitem{bian2019unsupervised}
J.~Bian, Z.~Li, N.~Wang, H.~Zhan, C.~Shen, M.-M. Cheng, and I.~Reid, ``Unsupervised scale-consistent depth and ego-motion learning from monocular video,'' \emph{Advances in neural information processing systems}, vol.~32, 2019.

\bibitem{paszke2019pytorch}
A.~Paszke, S.~Gross, F.~Massa, A.~Lerer, J.~Bradbury, G.~Chanan, T.~Killeen, Z.~Lin, N.~Gimelshein, L.~Antiga, \emph{et~al.}, ``Pytorch: An imperative style, high-performance deep learning library,'' \emph{Advances in neural information processing systems}, vol.~32, 2019.

\bibitem{kingma2014adam}
D.~P. Kingma and J.~Ba, ``Adam: A method for stochastic optimization,'' \emph{arXiv preprint arXiv:1412.6980}, 2014.

\bibitem{facil2019cam}
J.~M. Facil, B.~Ummenhofer, H.~Zhou, L.~Montesano, T.~Brox, and J.~Civera, ``Cam-convs: Camera-aware multi-scale convolutions for single-view depth,'' in \emph{Proceedings of the IEEE/CVF Conference on Computer Vision and Pattern Recognition}, 2019, pp. 11\,826--11\,835.

\bibitem{eigen2014depth}
D.~Eigen, C.~Puhrsch, and R.~Fergus, ``Depth map prediction from a single image using a multi-scale deep network,'' \emph{Advances in neural information processing systems}, vol.~27, 2014.

\bibitem{he2016deep}
K.~He, X.~Zhang, S.~Ren, and J.~Sun, ``Deep residual learning for image recognition,'' in \emph{Proceedings of the IEEE conference on computer vision and pattern recognition}, 2016, pp. 770--778.

\bibitem{yao2018mvsnet}
Y.~Yao, Z.~Luo, S.~Li, T.~Fang, and L.~Quan, ``Mvsnet: Depth inference for unstructured multi-view stereo,'' in \emph{Proceedings of the European conference on computer vision (ECCV)}, 2018, pp. 767--783.

\end{thebibliography}

\end{document}